\documentclass[letterpaper, 10 pt, conference]{ieeeconf}  %

\IEEEoverridecommandlockouts                              %

\usepackage{soul}
\usepackage[normalem]{ulem}
\usepackage[hidelinks]{hyperref}
\usepackage{bookmark}
\usepackage[latin1]{inputenc}
\usepackage[english]{babel}
\usepackage{mathtools}

\usepackage[dvipsnames]{xcolor}
\usepackage{amsfonts,amssymb,amsmath,color}
\usepackage{graphicx}
\usepackage{cite}
\usepackage{placeins}
\usepackage{graphicx}
\usepackage{algorithm}
\usepackage[noend]{algpseudocode}
\usepackage{todonotes}
\usepackage{balance}
\usepackage{tikz}
\usetikzlibrary{shapes,arrows,calc}
\usepackage{xpatch}
\newcommand\oprocendsymbol{\hbox{$\square$}}
\newcommand\oprocend{\relax\ifmmode\else\unskip\hfill\fi\oprocendsymbol}

\renewcommand{\theenumi}{(\roman{enumi})}

\newtheorem{theorem}{Theorem}[section]

\newtheorem{remark}[theorem]{Remark}

\def\QEDclosed{\mbox{\rule[0pt]{1.3ex}{1.3ex}}} %

\def\QED{\QEDclosed} %

\makeatletter
\newcommand{\StatexIndent}[1][3]{%
	\setlength\@tempdima{\algorithmicindent}%
	\Statex\hskip\dimexpr#1\@tempdima\relax}
\makeatother

\renewcommand{\lim}{\operatornamewithlimits{lim\vphantom{p}}}

\newcommand{\real}{{\mathbb{R}}}

\newcommand{\until}[1]{\{1,\ldots,#1\}}

\DeclareMathOperator{\col}{\text{col}}

\newcommand{\kron}{\otimes}

\newcommand\norm[1]{\lVert#1\rVert}

\newcommand{\agents}{V}
\newcommand{\intruders}{I}

\newcommand{\EE}{\mathcal{E}}
\newcommand{\Acal}{\mathcal{A}}

\newcommand{\GG}{\mathcal{G}}

\newcommand{\innbrs}{\mathcal{N}}

\newcommand{\n}{n}
\newcommand{\N}{N}

\usepackage{xparse}
\NewDocumentCommand{\xag}{O{}O{}}{x_{#1}^{#2}}
\NewDocumentCommand{\nag}{O{}}{\n_{#1}}
\NewDocumentCommand{\X}{O{}}{X_{#1}}
\NewDocumentCommand{\aij}{O{}}{a_{ij}^{#1}}
\NewDocumentCommand{\fag}{O{}O{}}{f_{#1}^{#2}}
\NewDocumentCommand{\hfag}{O{}O{}}{\hat{f}_{#1}^{#2}}
\NewDocumentCommand{\phiag}{O{}O{}}{\phi_{#1}^{#2}}

\newcommand{\xjneigh}{x_{\innbrs_i^t}}

\graphicspath{{figs/}}

\title{A Distributed Online Optimization Strategy \\
	for Cooperative Robotic Surveillance}

\author{%
	Lorenzo Pichierri, Guido Carnevale, Lorenzo Sforni, Andrea Testa and Giuseppe Notarstefano
	\thanks{This result is part of the project ``Distributed Optimization for Cooperative Machine Learning in Complex Networks" (No PGR10067) that has received funding from the Ministero degli Affari Esteri e della Cooperazione Internazionale.}%
	\thanks{Authors are with the Department of Electrical, 
		Electronic and Information Engineering, University of Bologna, Bologna, Italy.
		\texttt{\{lorenzo.pichierri, guido.carnevale, lorenzo.sforni, a.testa, giuseppe.notarstefano\}@unibo.it}.}%
}

\begin{document}

	\maketitle

	\begin{abstract} %
		In this paper, we propose a distributed algorithm to control a team of cooperating robots aiming to protect a target from a set of intruders. Specifically, we model the strategy of the defending team by means of an online optimization problem inspired by the emerging distributed aggregative framework. In particular, each defending robot determines its own position depending on (i) the relative position between an associated intruder and the target, (ii) its contribution to the barycenter of the team, and (iii) collisions to avoid with its teammates. We highlight that each agent is only aware of local, noisy measurements about the location of the associated intruder and the target. Thus, in each robot, our algorithm needs to (i) locally reconstruct global unavailable quantities and (ii) predict its current objective functions starting from the local measurements. The effectiveness of the proposed methodology is corroborated by simulations and experiments on a team of cooperating quadrotors.
	\end{abstract}

	\section{Introduction}
	\label{sec:introduction}
	The employment of autonomous mobile robots for security purposes is becoming more and more important, see e.g.~\cite{bullo2009distributed,rubio2019review,savkin2017distributed,marzoughi2021autonomous} and references therein. In this work, we investigate the framework in which a team of robots wants to protect a target (e.g., an asset or a region) from potential intruders.
	
	\paragraph*{Related Work}
	Several works address surveillance tasks in multi-robot settings by means of optimization procedures.
	Authors in~\cite{nigam2011control,natarajan2015multi,adaldo2017cooperative,petrlik2019coverage} model the surveillance task as a coverage control problem. 
	The work~\cite{simetti2010protecting} proposes an algorithm to protect an asset by optimizing the positioning of a team of autonomous vehicles. In~\cite{zhang2010decentralized}, authors propose a distributed, multi-objective algorithm to detect 
		intruders
		and protect sensitive areas.
		In~\cite{raboin2013model,mullins2015adversarial}, a team of autonomous vehicles is controlled to maximize the amount of time taken by an intruder to reach a protected asset. The work~\cite{popov2014robot} addresses the problem of target defense by a team of water vehicles represented by learning rhythmic motor primitives.
		In~\cite{acevedo2016distributed}, a dynamic, decentralized assignment algorithm based on one-to-one coordination is proposed to solve a multi-target allocation problem under communication constraints. 
		In~\cite{nantogma2019behavior}, a behavior-based fuzzy logic control system is presented to protect a target from potential threats. In~\cite{duan2020robotic}, a surveillance task is addressed by leveraging Markov chains.
	Recently, the emerging distributed aggregative optimization framework has gained attention to model scenarios arising in cooperative robotics. %
	In this field, a network of robots aims to minimize an objective function given by the sum of local functions which depend not only on a local decision variable (e.g., the position of the robot) but also on an aggregative quantity of the network (e.g., the barycenter of the team). This set-up has been introduced in the pioneering work~\cite{li2021distributed}. Constrained, online versions of the problem have been investigated in~\cite{li2021distributedOnline,carnevale2022distributed}. The authors of~\cite{chen2022distributed} consider communication with finite bits. In~\cite{carnevale2022aggregative}, a continuous-time distributed feedback optimization law is proposed to steer a set of single integrators to a steady-state configuration that is optimal with respect to an aggregative optimization problem. In~\cite{wang2022distributed}, a distributed algorithm based on the Franke-Wolfe update has been proposed to reduce the computational effort.
	
	\paragraph*{Contributions}
	Inspired by the seminal works in~\cite{li2021distributed,li2021distributedOnline}, in this paper we consider an online, 
	constrained optimization problem over peer-to-peer networks of robots. Differently from previous approaches, we propose
	a novel optimization paradigm in which local cost functions depend not only on their own optimization variables but also 
	on 
	optimization variables associated to other robots. Moreover, we consider a challenging scenario tailored for multi-robot
	settings in which robots have access to local functions only by means of noisy measurements. This requires the introduction
	of an ad-hoc estimation procedure. Finally, aiming at reducing
	the dynamic regret, we propose a novel resolution strategy requiring a prediction step on local, unavailable quantities.
	To the best of the authors' knowledge, this is the first work addressing such
	distributed setting. 
	This novel optimization paradigm is applied to a cooperative, 
	dynamic surveillance setting in which robots 
	have to 
	defend a target from a set of intruders while avoiding collisions. 
	Finally, we apply the proposed resolution strategy to a team of real aerial robots mimicking a basket match.
	
	\paragraph*{Organization} The paper unfolds as follows. In Section~\ref{sec:preliminaries} we introduce the dynamic multi-robot surveillance optimization problem.
	In Section~\ref{sec:modeling} we detail how to model the optimization problem
	to fulfill the surveillance task. 
	In Section~\ref{sec:algorithm} we introduce the distributed resolution strategies. 
	Simulations and experiments are provided in Section~\ref{sec:experiments}.
	
	\paragraph*{Notation}
	$I_n$ is the $n\!\times\! n$ identity. $\kron$ is the Kronecker product. $\text{blkdiag}(A,B)$ is the block diagonal concatenation of matrices $A$ and $B$. Given a symmetric, positive-definite matrix $Q\!\in\!\real^{n\!\times\! n}$, and $x\!\in\!\real^n$, we define the $Q$-norm of $x$ as $\Vert x \Vert_Q = \sqrt[2]{x^\top Q x}$.
		Given a function $g\! :\!\real^{n_1} \!\times\! \real^{n_2} \!\to \!\real$, $\nabla_1 g(\cdot,\cdot) \! \in \! \real^{n_1}$ and $\nabla_2 g(\cdot,\cdot) \!\in\! \real^{n_2}$ are the gradient of $g$ with respect to the first and the second argument, respectively.

	\section{Problem Formulation}
	\label{sec:preliminaries}
	In this section, we detail the distributed online surveillance problem for a multi-robot cooperating team. Before
	introducing the considered optimization framework, we introduce the communication model
	connecting the cooperating robots. 

	\subsection{Communication Model}
	In this paper, we consider a set $\agents = \until{N}$ of robots that have to accomplish a cooperative surveillance task. In this cooperative setting, robots have to leverage communications among each other in order to properly solve the problem. We assume that robots
	can exchange information according to a communication network modeled as a time-varying undirected graph
	$\GG^t=(\agents,\EE^t)$ 
	in which $\EE^t\subseteq \agents\times \agents$ is the edge set.
	A graph $\GG^t$ models the communication in the
	sense that there is an edge $(i,j), (j,i) \in \EE^t$ if and only if robots $i$ is able
	to exchange information with robots $j$ at time $t$. We consider $(i,i)\in\EE^t$ for each $t\geq 0$ and for each $i\in \agents$. 
	We assume that robots are able to communicate if their relative distance is below a certain
	threshold.
	For each node $i$, the set of \emph{neighbors}
	$j$ such that there exists an edge $(i,j) \in \EE^t$ of $i$ at time $t$ is denoted by 
	$\innbrs_i^t =\{j\in \agents \mid (i,j)\in\EE^t\}$.
	Given $B > 0$, a graph is said to be $B$-\emph{connected} if the graph $\{V,\bigcup_{\tau=t}^{t+B} \EE^{\tau}\}$ is connected.
	We also associate a so-called weighted adjacency matrix $\Acal^t \in \real^{N \times N}$ matching the graph, i.e., a matrix with $(i,j)$-entry $a_{ij}^t > 0$ if $(i,j) \in \EE$, otherwise $a_{ij}^t = 0$. The adjacency matrix is said to be doubly stochastic if $\sum_{i=1}^N a_{ij}^t = 1$ for all $j\in\until{N}$ and $\sum_{j=1}^N a_{ij}^t =1 $ for all $i\in\until{N}$.
	
	In the next, we assume that, for all $t \ge 0$, the communication graph $\GG^t$ is $B$-connected and $\Acal^t$ is doubly stochastic. Moreover, we also assume that there exists $a \in (0,1)$ such that, for all $t \ge 0$ and $i,j \in \until{N}$, it holds $a_{ij}^t > a$ whenever $a_{ij}^t > 0$.
	
	\subsection{Distributed Online Optimization for Robotic Surveillance}

	The cooperative task consists of the surveillance of a certain, possible moving target. 
	More in detail, we consider a dynamic scenario in which a team $\intruders=\until{\N}$ of intruding, adversarial robots moves towards the target
		(see, e.g., target surrounding problem in \cite{li2021distributedOnline,li2021distributed})
	.
	Robots are modeled as generic discrete-time nonlinear systems.
	The generic robot $i\in \agents$ is aware of only one of the attackers $i\in \intruders$.
	Moreover, it does not have any other information about other robots in $\agents$ and $\intruders$.
	To fulfill this defensive task,
	at the generic time $t\geq 0$  each robot in $i \in \agents$, cooperating with its teammates, chooses its configuration
	$\xag[i][t]\in\real^3$ in the space according to the following policies:
	\begin{enumerate}
		\item  stay between the attacker $i \in \intruders$ and the target;
		\item  stay near to other teammates and steer the barycenter of the team towards the target;
		\item  avoid collisions with other robots in $\agents$;
		\item  move in a bounded area of the field.
	\end{enumerate}
	
\begin{figure}[h!]
		\centering
		\includegraphics[width=.7\columnwidth]{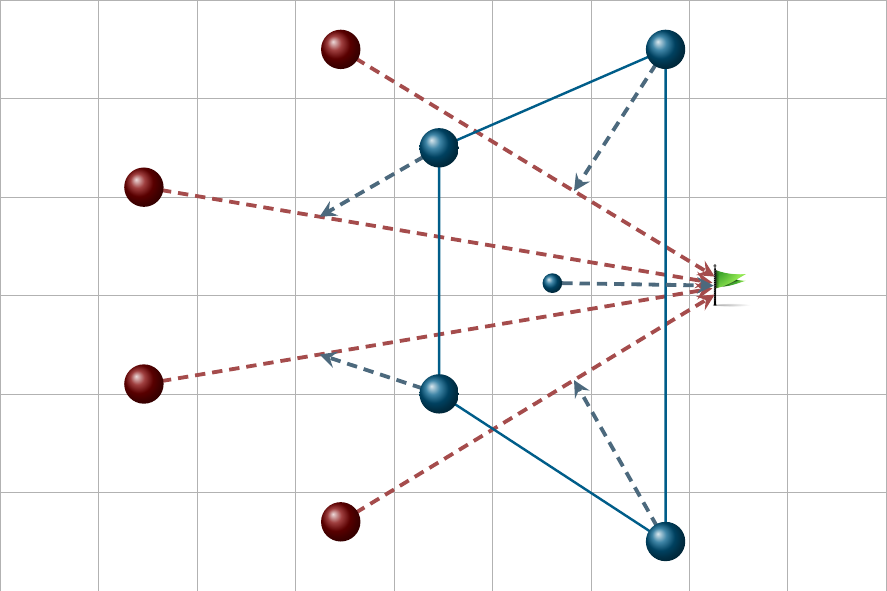}
		\caption{Example of desired behavior. Each robot $i\in \agents$ wants to move in the line connecting the attacker $i\in \intruders$ with the target (green flag). The barycenter (small blue sphere) is steered toward the flag.}
		\label{fig:surveillance}
	\end{figure}

	An illustrative example of this set-up is provided in Figure~\ref{fig:surveillance}. Here, each member of the defending team $\agents$, represented by blue spheres, wants to move between the designated intruder and the target, represented as a flag. Moreover, the team of robots $\agents$ is steering its barycenter towards the target.

	We want to find the optimal configuration of the defending team with respect to the aforementioned tasks.
	To this end, we model this set-up by means of a distributed online optimization problem taking inspiration from the aggregative optimization framework~\cite{li2021distributed,li2021distributedOnline,carnevale2022distributed,chen2022distributed,carnevale2022aggregative,wang2022distributed}.
	In this optimization set-up, robots  in $V$, at the generic time $t \ge 0$, aim at solving the following problem
	\begin{align}
		\begin{split}
			\min_{\xag\in \X^t} \: & \: \sum_{i =1}^\N \fag[i][t](\xag[i],\sigma(\xag),\xjneigh).
		\end{split}
		\label{eq:aggregative_problem}
	\end{align}
	Here, $\xag = \col(\xag[1],\ldots,\xag[\N])$, $\xjneigh=\{\xag[j]\}_{j\in\innbrs_i^t\setminus \{i\}}$, each $\fag[i][t](\xag[i],\sigma(\xag),\xjneigh): \real^3 \times \real^3\times\real^{3(|\innbrs_i^t|-1)}   \mapsto\real$
	depends on the optimization variables of a subset of robots in $\agents$ and on an \emph{aggregative variable}
	$\sigma(\xag) \triangleq \sum_{i=1}^\N \frac{\phiag[i](\xag[i])}{\N}$, 
	with $\phiag[i](\xag[i]):\real^{3}\mapsto \real^3$ for each $i\in\until{\N}$.
	We denote by $\X^t=\prod_{i=1}^\N \X[i]^t$ a constraint set on the position of robots at time $t$, where $\X[i]^t \subseteq \real^{3}$.
	The overall objective function is %
	\begin{align*}
		f^t(\xag,\sigma(\xag))=\sum_{i =1}^\N f_{i}^t(\xag[i],\sigma(\xag),\xjneigh).
	\end{align*}
	In the next section, we provide guidelines on how to shape the functions $f_{i}^t,\phiag[i]$, and the sets $\X[i]^t$ in order to affect the defensive behavior.
	
	Throughout this paper, we consider a cooperative, distributed scenario in which robots have limited knowledge of the problem~\eqref{eq:aggregative_problem}. Thus, in order to solve the optimization problem, they have to iteratively exchange suitable information via a communication network.
	In classical distributed online optimization frameworks~\cite{li2022survey}, each robot $i\in \agents$ can only privately access $\fag[i][t],\phiag[i]$, and $\X[i]^t$ only once $\xag[i][t]$ has been computed. %
	We instead consider a challenging scenario in which the robots in $\agents$ have to estimate these quantities according to noisy measurements on the attackers and on the target. 
	Moreover, in our setup, the quantities $\fag[i][t]$ and $\X[i]^{t}$ are predicted before evaluating the optimal location estimate $\xag[i][t]$.
	\section{Modeling}
	\label{sec:modeling}
	In this section, we detail a strategy to model the defensive behavior of the robotic team $\agents$ within the considered distributed online optimization setting. In the first part, we show how to model the cost function of the generic robot to surveil a certain adversary and how to model team behaviors for target protection and collision avoidance. In the second part of the section, we show how to properly constrain the movements of each robot.
	Finally, we provide details on how to predict the movements of the intruders.
	We do not consider any knowledge of the behavior of the offending team.

	\subsection{Modeling of Cost Function}
	\label{sec:task12}
	In this section, we model the objective function of the generic robot $i$ as in~\eqref{eq:aggregative_problem} as the sum of three terms, i.e.,
		$\fag[i][t](\xag[i],\sigma(\xag),\xjneigh)\!\!=\!\!\fag[iS][t](\xag[i])+\fag[iC][t](\xag[i],\sigma(\xag))+
		\fag[iB][t](\xag[i],\xjneigh)$.
	More in detail, the first term $\fag[iS][t](\xag[i])$ models the behavior of the single robot and solely depends on the position $\xag[i]$ of the robot. The second term $\fag[iC][t]$ instead models a collective behavior and depends also on the position of other robots in the team $\agents$. 
	The third term $\fag[iB][t](\xag[i],\xjneigh)$ instead handles collisions with other robots in $\agents$.
	We recall that the $i^{\text{th}}$ robot does not have any knowledge of other robot positions. Moreover, robot $i$ is only able to measure the position $p_i^t$ of an intruder $i\in \intruders$ at time $t$, thus not having knowledge of its future movements.
	The solution strategy detailed in Section~\ref{sec:algorithm} will handle this lack of knowledge.

	Each defending robot wants to track the associated offending intruder.
	Such defending strategy can be captured by defining for each robot $i$ a
	local cost $\fag[iS][t]$ in the form
	$\fag[iS][t](\xag[i]) = \Vert  \xag[i]-p_{i}^{t} \Vert_{Q_{i1}}^2$,
	where $Q_{i1} \in \real^{3\times 3}$ is a symmetric, positive definite matrix. In this way, the robot is steered towards the adversary player. 
	
	In practical scenarios, it is desirable that the barycenter of the team $\agents$ is near the target while robots surveil intruders. Moreover, robots may not want to scatter in the environment. Rather, they may want to be close to each other.  Let $b^t \in\real^3$ be the position of the target at time $t$.
	Such defending strategy can be obtained by defining a local function $\fag[iC][t]$
	of the form
	$\fag[iC][t](x_i,\sigma(x)) = \Vert \sigma(x) - b^t \Vert_{Q_{i2}}^2 + \Vert \sigma(x) - \xag[i] \Vert_{Q_{i3}}^2$,
	where $Q_{i2}, Q_{i3}\in\real^{3\times 3}$ are symmetric, positive definite matrices, while $\sigma(x)$ is designed to represent the center of mass of robots in $\agents$, i.e., $\sigma(x) = \frac{\sum_{i=1}^\N \xag[i]}{\N}$.

	In order to accomplish the collision avoidance task, the third term in the objective function is in the form 
	$\fag[iB][t](\xag[i],\xjneigh) = \sum_{j\in\ \innbrs_i^t\setminus\{i\}} \mathcal{B}(\xag[i],\xag[j])$ where $\mathcal(\cdot,\cdot)$ is some barrier function. 
	A suitable example is
	\begin{align}
		\fag[iB][t](\xag[i],\xjneigh) = \sum_{j\in \innbrs_i^t\setminus\{i\}} -\log(\Vert\xag[i]-\xag[j]\Vert).\label{eq:barrier}
	\end{align}
	In practical settings, the generic robot $i$ is not required to know the position $\xag[j]$ of other robots. Indeed, the distance $\Vert\xag[i]-\xag[j]\Vert$ and the relative distance vector $\xag[i]-\xag[j]$ can be retrieved, e.g. using Lidar sensors. Thereby, the function and its gradients can be evaluated using suitable measurements.

	\subsection{Modeling of Constraint Set}
	\label{sec:task3}
\begin{figure}[h!]
		\centering
		\includegraphics[width=.7\columnwidth]{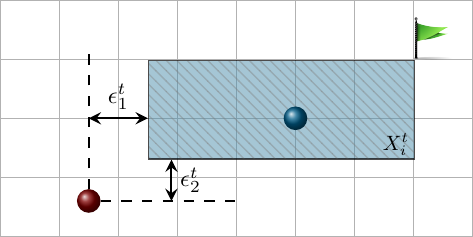}
		\caption{Example of the constraint set for one of the robots $i\in \agents$. Robot $i$ (blue sphere), can move in the constraint set as in~\eqref{eq:basket_con_2} between the adversary (red sphere) and the target (green flag).} \label{fig:constr}
	\end{figure}

	Let us introduce the constraints. We define $X_{\text{field}}\subseteq\real^3$ as the set
	of positions contained within the operating region. %
	In order to protect the target, we impose that each robot always stays between the intruder and the target, i.e., it must satisfy
	\begin{align}
		\begin{cases}
			[p_{i}^t]_c + \epsilon_c^t \le [\xag[i][t]]_c \le [b^t]_c  & \! \text{if } [p_{i}^t]_c \le [b^t]_c 
			\\
			[b^t]_c \le [\xag[i][t]]_c \le [p_{i}^t]_c  - \epsilon_c^t& \! \text{if } [p_{i}^t]_c > [b^t]_c 
		\end{cases} \quad \!\!\!\!\!\!c =1, 2, 3,
		\label{eq:basket_con_2}
	\end{align}
	where $[\cdot]_c$ denotes the $c$-th component of the vectors and $\epsilon_c^t = \max(\epsilon_{c,\text{min}},\kappa_c |[p_{i}^t]_c - [b^t]_c |^2)$ is an additional tolerance with $\kappa_c > 0$ and a saturation term $\epsilon_{c,\text{min}} > 0$. The overall feasible set $\X[i]^t$ of each
	robot $i$ at time $t$ is then
	\begin{align}
		\X[i]^t = X_{\text{field}} \cap \{\xag[i] \in \real^3 \mid \eqref{eq:basket_con_2} \text{ is satisfied} \}.
	\end{align}
	We explain as follows the rationale behind the constraint sets $\X[i]^t$. Each defending robot $i \in \agents$ wants to stay between the corresponding intruder $i \in \intruders$ and the target to be protected. The additional tolerances $\epsilon_c^t$ provide an operating margin that can be used by the defending robots to manage the fact that offending robots (and, possibly, the target) move over time. In particular, these tolerances are as small as the distance between the intruders and the target is small. Indeed, it is rather intuitive the fact that the defending robots need to stay closer to the intruders when they are close to the target. 
	An illustrative 2D example of this constraint set is in Figure~\ref{fig:constr}.

	\section{Solution Strategy}
	\label{sec:algorithm}
	
	\subsection{Prediction of Intruder and Target Behavior}
	\label{sec:predictor}
	
	As we detail in the next part of this section, in the proposed distributed strategy we 
	rely on a
	prediction of target and adversarial moves.
	At the same time, 
	since the precise intruder (or target) position
	could not be available at each time instant, e.g. the intruder is out of sight, 
	it is necessary to determine the unknown position just by considering
	the preceding history.
	To overcome both these issues, 
	and assuming sufficiently slow dynamics,
	we adopted a Kalman filter-based solution,~\cite{kalman1960new}
	.

	For the sake of simplicity, we model each adversarial robot as a 
	discrete-time double integrator system.
	Since we have no access to the input applied to each system, 
	we approximate each member of the adversarial team as 
	an autonomous system driven by a normally distributed signal, i.e.,
	\begin{align}
		\label{eq:kf:sys}
		\begin{bmatrix}
			p_{i}^{t+1} 
			\\
			v_{i}^{t+1}
		\end{bmatrix}
		= 
		\underbrace{\begin{bmatrix}
				I_3 & \Delta t \; I_3
				\\
				0 & I_3
		\end{bmatrix}}_{=: F}
		\begin{bmatrix}
			p_{i}^{t} 
			\\
			v_{i}^{t}
		\end{bmatrix}
		+ 
		\underbrace{\begin{bmatrix}
				0
				\\
				\Delta t \; I_3
		\end{bmatrix}}_{=: G}
		a_i^t
	\end{align}
	where $v_{i}^t \in \real^3$ is the velocity of the intruder $i$ at time instant $t$,
	$\Delta t \in \real$ the discretization step and $a_i^t \in \real^3$ represents the effects of 
	the unknown input applied to the $i^{\text{th}}$ system. Specifically, we assume $a_i^t$ drawn from a 
	normal distribution with zero mean and covariance matrix $\sigma_i^2 I$.
	Let $H$ be defined as $H:= [I_3 \;\; 0]$. We assume we can access some measurements of the adversaries' states, namely,
	at each time $t$, we observe 
		\begin{align}
			\label{eq:kf:meas}
			z_{i,p}^t = H [(p_{i}^{t})^\top, (v_{i}^{t})^\top]^\top
			+
			w_{i,p}^t
	\end{align}
	where $z_{i,p}^t \in \real^3$ is the output of the $i^{\text{th}}$ system and $w_{i,p}^t$ is a normally distributed
	disturbance with zero mean and covariance matrix $R_{i,p}$. 
	Firstly, let us introduce some notation. We denote as $\xi_{i,p}^t$ the state of each system $i$ at time $t$, i.e., $\xi_{i,p}^t := \col(p_i^t, v_i^t)$. The estimation, computed at time $t$, of $\xi_{i,p}^{t+1}$ is denoted as $\hat{\xi}_{i,p}^{t+1}$. Moreover, to each estimate $\hat{\xi}_{i,p}^{t+1}$ we can associate a covariance matrix $P_{i,p}^{t+1} \in \real^{6 \times 6}$, which can be seen as a measure of the estimated accuracy of the current state estimate. 
	We now recall the Kalman filter equations applied to our setting.
	Specifically, the approach is an iterative process consisting of two phases: prediction 
	and correction.
	For all $t$, in the prediction phase a new state estimation (and covariance matrix), based on the system model is computed, namely,
		\begin{align}
			\hat{\xi}_{i,p}^{t+1} = F \hat{\xi}_{i,p}^{t} 
			\qquad
			P_{i,p}^{t+1} = F P_{i,p}^{t} F^\top + S_{i,p}
			\label{eq:kf:prediction}
		\end{align}
	where $S_{i,p} := G G^\top \sigma_i^2$. 
	As for the correction phase, as soon as a measurement $z_{i,p}^{t}$ of $\xi_{i,p}^{t}$ is available (via~\eqref{eq:kf:meas}), we correct the estimation obtained by~\eqref{eq:kf:prediction} via
	\begin{subequations}
		\label{eq:kf:correction}
		\begin{align}
			\hat{\xi}_{i,p}^{t+1} &= \hat{\xi}_{i,p}^{t+1} + K_i(z_i^{t} - H \hat{\xi}_{i,p}^{t})
			\\
			\hspace*{-0.5cm}
			P_{i,p}^{t+1} &= (I_6 - K_i H)P_{i,p}^{t+1}(I_6 - K_i H)^\top + K_i R_{i,p} K_i^\top
		\end{align}
		where $K_i$ is the so-called Kalman gain computed as
		\begin{align}
			K_{i,p} = F P_{i,p}^t H^\top(H P_{i,p}^{t} H^\top + R_{i,p})^{-1}.
		\end{align}
	\end{subequations}
	The updates~\eqref{eq:kf:prediction} and~\eqref{eq:kf:correction} are performed iteratively 
	with initial conditions $ \xi_{i,p}^{0} = \xi_{i,p}^{\text{init}}$, 
	and $P_{i,p}^{0} = P_{i,p}^{\text{init}}$. In general, $P_{i,p}^{\text{init}}$ can be seen as a measure of the accuracy of the initial state estimate $\xi_{i,p}^{\text{init}}$.
	The prediction-correction update can be written in its compact form 
	\begin{align*}
		\hat{\xi}_{i,p}^{t+1} &= (F - K_{i,p} H)\hat{\xi}_{i,p}^{t}  + K_{i,p} z_{i,p}^{t} 
		\\
		P_{i,p}^{t+1} &= (F\!\! -\!\! K_{i,p} H)P_{i,p}^{t}(F \!\! - \!\! K_{i,p} H)^\top\!\! + \!\!K_{i,p} R_{i,p} K_{i,p}^\top + S_{i,p}
	\end{align*}
	At time $t$ the intruder position can be estimated as
	$\hat{p}_i^t = H \hat{\xi}^t$.
	A similar approach is adopted to estimate the target position. That is, target dynamics is approximated as (cf.~\eqref{eq:kf:sys}),
	$\xi_{b}^{t+1} = F \xi_{b}^{t} + G a_b^t$,
	where $\xi_{b}^{t} := \col(b^t, v_b^t)$
	with $b^t$ is the position of the target and $v_b^t$ its velocity and
	$a_b^t$ is sampled from a normal random variable with covariance matrix $\sigma_b^2 I$.
	Robot $i$ accesses the noisy measurement 
	$z_{i,b}^t = H \xi_{b}^t
	+
	w_{b,i}^t$ of $\xi_{b}^{t}$. The disturbance
	$w_{b,i}^t$ is normally distributed with zero mean and covariance matrix $R_{b,i}$. 
	Each robot then computes an estimate $\hat{\xi}_{i,b}^t$ of $\xi_{b}^t$,
	with its covariance $P_{i,b}^t$,
	as in~\eqref{eq:kf:prediction}-\eqref{eq:kf:correction}.
	Hence, the target position estimate $\hat{b}_i^t$ computed by agent $i$ can be
	obtained similarly as $\hat{b}_i^t = H\hat{\xi}_b^t$.
	The overall procedure is implemented by each robot via Algorithm~\ref{table:constrained_aggregative_GT}
	where the following notation is adopted. 
	Let us define $\bar{F} := I_2 \kron F$, $\bar{H} := I_2 \kron H$, $S_{i} := \text{blkdiag}(S_{i,p}, S_{i,b})$, $R_i := \text{blkdiag}(R_{i,p}, R_{i,b})$, $K_i := \text{blkdiag}(K_{i,p}, K_{i,b})$.
	Moreover, we denote as $P_i^t := \text{blkdiag}(P_{i,p}^t, P_{i,b}^t)$, $\hat{\xi}_i^t := \text{col}(\hat{\xi}_{i,p}^t, \hat{\xi}_{i,b}^t)$, $z_i^t := \text{col}(z_{i,p}^t, z_{i,b}^t)$ and $w_i^t := \text{col}(w_{i,p}^t, w_{i,b}^t)$.

	\subsection{Distributed Prediction and Optimization Scheme}
	
	The defending team runs Algorithm~\ref{table:constrained_aggregative_GT} 
	(see the table below)
	to choose the positions $\xag[i][t]$ for all $t \ge 0$. The aim of the algorithm is to minimize the dynamic regret $R_T$, i.e., the performance metric
		$R_T := \textstyle\sum_{t=1}^T (f^t(\xag[][t],\sigma(\xag[][t])) - f^t(x^t_\star,\sigma(x_\star^t)))$,
	where $T > 1$ denotes a given time horizon and $x^t_\star \in \real^3$ denotes a minimizer of $\fag[][t](\cdot,\sigma(\cdot))$ over the set $\X^t$. We underline that the objective functions introduced in Section~\ref{sec:task12} are strongly convex, and the constraint set described in Section~\ref{sec:task3} is non-empty, closed, and convex. Thus, the minimizer $x^t_\star$ is unique for all $t$.
	
	The algorithm structure described in Algorithm~\ref{table:constrained_aggregative_GT} relies on two main features. 
	The first one regards a (local) tracking mechanism that allows each robot in the network to reconstruct global quantities that cannot be directly accessed.
	The second feature involves a (local) prediction mechanism giving rise to the current estimates of both the (local) objective function and the feasible set. 
	Indeed, as usual in the context of distributed online optimization (see, e.g., the recent survey~\cite{li2022survey}), we assume that both $\fag[i][t+1]$ and $\X[i]^{t+1}$ are revealed to robot $i$ only once the update $\xag[i][t+1]$ has been computed. 
	As a consequence, even with strongly convex objective functions, the existing bounds on the dynamic regret necessarily suffer the presence of terms $O(\sum_{t=1}^{T-1}\norm{\xag[\star][t+1] - \xag[\star][t]})$,~\cite{li2022survey}. 
	We overcome this issue by implementing the estimation and prediction mechanism in Section~\ref{sec:predictor}, so that each robot $i$ of the network is able to compute the estimates $\hfag[i][t+1]$ and $\hat{X}^{t+1}_{i}$ of the objective function $\fag[i][t+1]$ and the feasible set $\X[i]^{t+1}$, respectively. We highlight that such a mechanism can be used thanks to the specific structure of the problem modeled in Section~\ref{sec:modeling}. Indeed, 
	both $\fag[i][t+1]$ and $\X[i]^{t+1}$ 
	could be fully computed assuming $p_{i}^{t+1}$ and $b^{t+1}$ known at time $t$ and approximating $\fag[iB][t+1]$ with $\fag[iB][t]$.

	We now depict the main idea behind the steps \eqref{eq:feasible_direction}--\eqref{eq:y_local_update} of Algorithm~\ref{table:constrained_aggregative_GT}. 
	In order to solve problem~\eqref{eq:aggregative_problem}, one may implement a projected gradient method in a distributed way (see, e.g.,~\cite{bertsekas2016nonlinear}). Indeed, when applied to problem~\eqref{eq:aggregative_problem}, the $i^{\text{th}}$ block should read
	\begin{align}\label{eq:desired_update}
		\xag[i][t+1] = P_{\X[i]^{t+1}}\left[\xag[i][t] - \alpha[\nabla \fag[][t+1](\xag[][t],\sigma(\xag[][t]))]_i\right],
	\end{align}
	where $P_{\X[i]^{t+1}}[\cdot] \in \real^3$ denotes the projection operator over the set $\X[i]^{t+1} \subseteq \real^3$ and, with slight abuse of notation, $[v]_i \in \real^3$ denotes the $i^{\text{th}}$ block in $\real^3$ of a given vector $v \in \real^{3N}$. 
	However, we remark that the locations $p_{i}^{t+1}$ and $b^{t+1}$ are measured only once $\xag[i][t+1]$ has been computed and, hence, we can only use the predictions $\hat{p}_i^{t+1}$ and $\hat{b}_i^{t+1}$ to obtain $\hfag[][t+1]$ and $\hat{X}_{i}^{t+1}$. Hence, we rewrite~\eqref{eq:desired_update} as
	\begin{align*}
		\xag[i][t+1] = P_{\hat{X}_i^{t+1}}\left[\xag[i][t] - \alpha[\nabla \hfag[][t+1](\xag[][t],\sigma(\xag[][t]))]_i\right].
	\end{align*}
	Now, by applying the chain rule 
	we observe that 
		\begin{align*}
			[\nabla \hfag[][t+1](\xag[][t],\sigma(\xag[][t]))]_i 
			&= \nabla\hfag[iS][t+1](\xag[i][t])
			+\nabla_1\hfag[iC][t+1](\xag[i][t],\sigma(\xag[][t]))
			\notag\\
			& 
			+ \nabla\phiag[i](\xag[i][t])\frac{1}{N}\sum_{j=1}^N \nabla_2 \hfag[jC][t+1](\xag[j][t],\sigma(\xag[][t]))
			\notag\\
			&
			+ 2\nabla_1\fag[iB][t](\xag[i][t],\xjneigh^t).
		\end{align*}
	However, as highlighted above, both quantities $\sigma(\xag[][t])$ and $\sum_{j=1}^N\nabla_2 \hfag[jC][t+1](\xag[j][t],\sigma(\xag[][t]))$ are global information that, in our setting, cannot be locally accessed. In order to compensate for this lack of knowledge, we introduce two auxiliary local variables $s_{i}^t, y_i^t \in \real^3$ called trackers. As shown in~\eqref{eq:s_local_update} and~\eqref{eq:y_local_update}, both the trackers are updated according to a perturbed consensus dynamics,~\cite{kia2019tutorial}. 
		Finally, in~\eqref{eq:convex_combination} a convex combination step is performed.
		\begin{remark}
			The steps of Algorithm~\eqref{eq:local_algorithm} resemble the ones of the scheme in~\cite{carnevale2022distributed}, where a slightly different version of problem~\eqref{eq:aggregative_problem} without collision avoidance functions is considered.
			In~\cite{carnevale2022distributed}, in the case of strongly convex problems, the following results are provided: (i) an upper bound for the achieved dynamic regret, and (ii) linear convergence in the static setup. 
			Moreover, the updates $x_{i}^{t+1}$ in~\cite{carnevale2022distributed}, only use (old) information about $f^t_i$ without any prediction, as we do instead in the proposed scheme.
		\end{remark}

	\begin{algorithm}%
		\begin{algorithmic}
			\State initialization:
			\begin{align*}
				x_{i}^0 &\in X_{i}^0, & s_{i}^0 &= \phiag[i](\xag[i][0]), & y_{i}^0 &= \nabla_2f_{i}^0(x_{i}^0,s_{i}^0) 
				\\
				\hat{\xi}_i^0 &= \col(\xi_{i,p}^0,\xi_{i,b}^0) & P_i^0 &= 0_{6\times6}
			\end{align*}
			\For{$t=0, 1, \dots$}

				\State Measure $\: z_i^t = \bar{H} \xi_i^t + w_i^t$
			\State Predict
			\begin{align*}
				\hat{\xi}_{i}^{t+1} &= (\bar{F} - K_i \bar{H})\hat{\xi}_{i}^{t}  + K_i z_{i}^{t} 
				\\
				P_{i}^{t+1} &= (\bar{F} - K_i \bar{H})P_{i}^{t}(\bar{F} - K_i \bar{H})^\top + K_i R_{i} K_i^\top + S_{i}
				\\
				(\hfag[i][t+1]&,\hat{\X}_{i}^{t+1}) \leftarrow \hat{\xi}_{i}^{t+1}
			\end{align*}
			\vspace{-6mm}
			\State Optimize
			\vspace{-2mm}
			\begin{subequations}\label{eq:local_algorithm}
				\begin{align}
					&\tilde{x}_{i}^{t} \!=\! P_{X_{i}^{t}} \big[x_{i}^{t}\! - \!\alpha(\nabla_1 \hfag[i][t](x_{i}^{t},s_{i}^{t},\xjneigh^t)\! + \!\nabla\phiag[i](x_{i}^{t})y_{i}^{t})\big]
					\label{eq:feasible_direction}
					\\
					&x_{i}^{t+1} \!=\! x_{i}^{t} + \delta(\tilde{x}_{i}^{t} - x_{i}^{t})\label{eq:convex_combination}
					\\
					&s_{i}^{t+1} \!=\! \textstyle\sum_{j \in \innbrs_i^t}a_{ij}^ts_{j}^{t} + \phiag[i](x_{i}^{t+1}) - \phiag[i](x_{i}^{t})\label{eq:s_local_update}
					\\
					&y_{i}^{t+1} \!=\! \textstyle\sum_{j \in \innbrs_i^t}a_{ij}^t y_{j}^{t} \! + \! \nabla_2 \hfag[i][t+1](x_{i}^{t+1}, s_{i}^{t+1},\xjneigh^{t+1})
					\notag\\
					&\hspace{2.5cm} - \nabla_2 \hfag[i][t](x_{i}^{t},s_{i}^{t},\xjneigh^t)\label{eq:y_local_update}
				\end{align}
				\vspace{-5mm}
			\end{subequations}
			\EndFor
		\end{algorithmic}
		\caption{From the perspective of robot $i$}
		\label{table:constrained_aggregative_GT}
	\end{algorithm}
	Notice also that since the constraint set is designed according to Section~\ref{sec:task3}, then the projection step in~\eqref{eq:feasible_direction} can be performed by thresholding each component.
	As for the communication burden, steps \eqref{eq:s_local_update}-\eqref{eq:y_local_update} require robots to exchange at each communication round two vectors in $\real^3$. 
		Thereby, the exchanged data always consists of $6$ floats.

	\section{Simulative and Experimental Results}
	\label{sec:experiments}
	
	In this section, we validate the proposed strategy in both a surveillance scenario (via simulations)
	and a basketball-like setting (via experiments). 
	All the simulations and experiments are performed using \textsc{ChoiRbot}~\cite{testa2021choirbot},
	a novel ROS~2 framework tailored for multi-robot applications. 
	Specifically, each robot is controlled by a set of independent ROS processes
	implementing the distributed algorithm. To this end, processes leverage the 
	DISROPT~\cite{farina2020disropt} functionalities. 
	In ROS~2, inter-process communication is implemented via the TCP/IP stack.
	This allows us to implement distributed algorithms, both in simulations and experiments, 
	on a real WiFi network. Thus,
	each robot only communicates with a few neighbors according to a given graph.
	In both the simulations and the experiment, we adopt Algorithm~\ref{table:constrained_aggregative_GT}, with $\alpha = 0.2$ and $\delta = 0.4$.
	As for the Kalman filter, we consider $\Delta t = 0.01$ s, $S_i = 10\,I_{12}$ and $R_i = 10^{-4}\, I_4$, for all $i \in \agents$.

	\subsection{Surveillance strategies}
	\label{sec:surveillance_strategies}
	
\begin{figure}[h]
			\centering
			\includegraphics[width=.45\columnwidth]{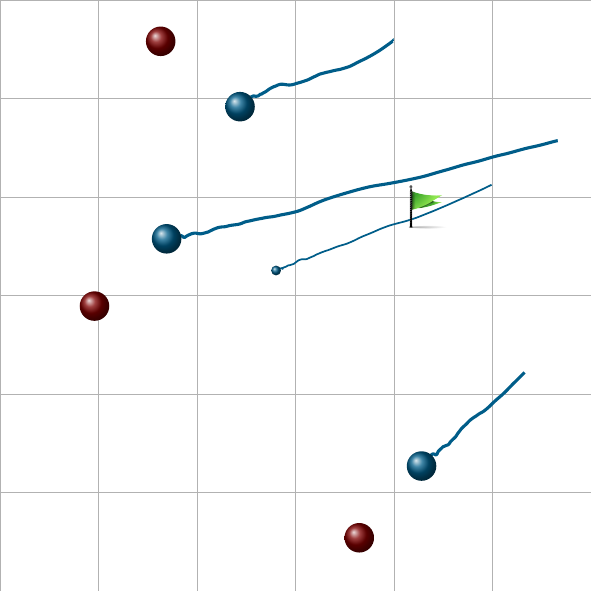}
			\hspace{1mm}
			\includegraphics[width=.45\columnwidth]{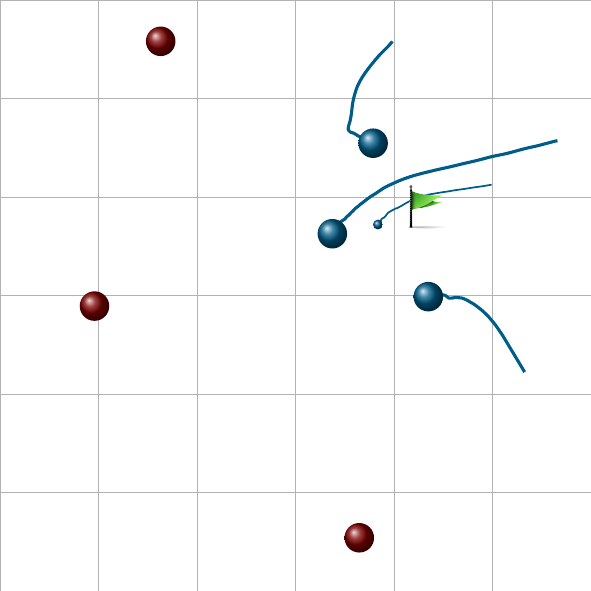}
			\caption{Top-down view of simulation. Blue spheres denote robots in $\agents$, 
					red spheres robots in $\intruders$, green flag denotes the target.
					We pick $\gamma_{i,p} = 10.0$, $\gamma_{i,b} = 5.0$, $\gamma_{i,\text{agg}} = 0.1$
					in both simulations, $\lambda_i = 0.8$ left, 
					$\lambda_i = 0.2$ right. %
			}
			\label{fig:surveillance_pressing}
		\end{figure}

\begin{figure}[h]
			\centering
			\includegraphics[width=.45\columnwidth]{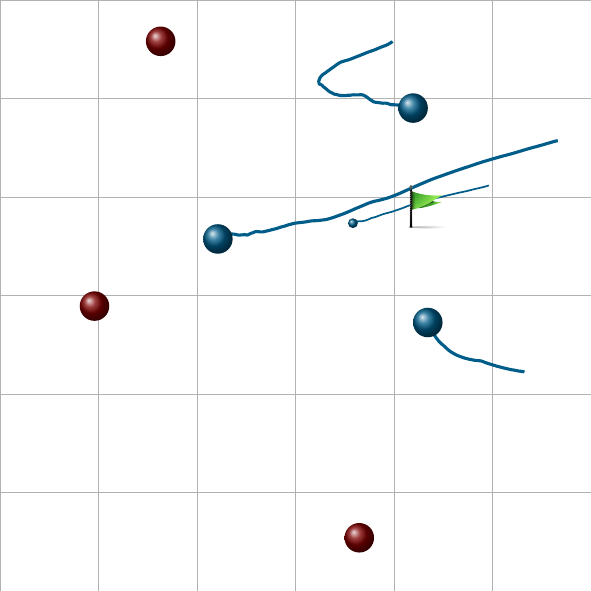}
			\hspace{1mm}
			\includegraphics[width=.45\columnwidth]{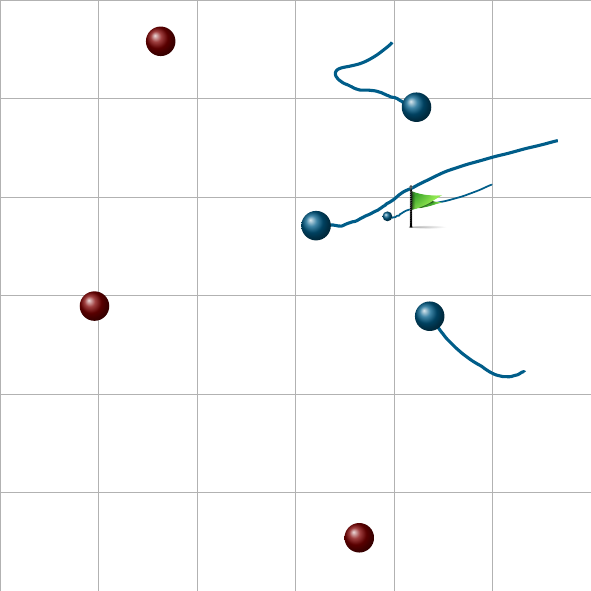}
			\caption{
					Here, $\lambda_1 = 0.5, \lambda_2 = 0.8, \lambda_3 = 0.2$ and $\gamma_{i,p} = 2.0$
					in both simulations, $\gamma_{i,b} = 5.0, \gamma_{\text{agg}} = 5.0$ (left), 
					$\gamma_{i,b} = 20.0, \gamma_{\text{agg}} = 20.0$ (right)}. %
			\label{fig:surveillance_aggregation}
		\end{figure}

	In this subsection, we consider a static scenario where 
	a defender team of $N = 3$
	robots 
	aim to protect a (fixed) target from an intruder squad with $3$ (static) members. 
	For the sake of simplicity, we consider all the robots moving at a fixed height.
	We model the surveillance 
	strategy 
	by solving a static instance of problem~\eqref{eq:aggregative_problem} with costs
		\vspace*{-2mm}
		\begin{align*}
			\fag[i][t]&(\xag[i],\sigma(\xag),\xjneigh) =
			\gamma_{i,p}\norm{x_i - \tilde{p}_i}^2 + 
			\gamma_{\text{i,agg}}\norm{\sigma(x) - b}^2 +
			\\
			&+\gamma_{i,b}\norm{\sigma(x) - x_i}^2 
			+\sum_{j\in \innbrs_i^t} -\log(\Vert\xag[i]-\xag[j]\Vert),
		\end{align*}
	for all $i \in \agents$, with $\gamma_{i,p}, \gamma_{\text{agg}} > 0$,
	$\tilde{p}_i \in \real^3$ denotes a point lying on the segment connecting the $i^{\text{th}}$ 
	intruder with the target, i.e., $\tilde{p}_i = \lambda_i p_i + (1-\lambda_i)b$ with $\lambda_{i} \in [0,1]$.
	The more the $i^{\text{th}}$ 
	intruder is considered dangerous, the more the defensive strategy is aggressive (i.e., $\lambda_i\to 1$). 
		As a consequence, the $i^{\text{th}}$ defender positions itself closer to the corresponding intruder.
	The impact of this parameter is shown in 
	Figure~\ref{fig:surveillance_pressing}. 
	Specifically, 
		an empirical tuning shows that,
		Figure~\ref{fig:surveillance_pressing} (left), 
		increasing the parameters $\lambda_i$, 
		the final configuration of the robots in $\agents$ results closer to the robots in $\intruders$. 
	Instead in Figure~\ref{fig:surveillance_pressing} (right), we reduce the magnitude of these coefficients and the defenders place closer to the target.
	The parameters $\gamma_{i,b}$ are meant to enforce 
	the defenders to keep their barycenter as close as possible to the target.
	The coefficients $\gamma_{i, \text{agg}}$ are used to avoid the robots to be 
	scattered away from each other.
	This scenario is depicted in Figure~\ref{fig:surveillance_aggregation}.
	In particular, in the right picture, we boost both $\gamma_{i,b}$ and $\gamma_{\text{i,agg}}$, to highlight these effects.
	\subsection{Multi-Robot Basketball Scenario}
	\label{sec:basket}
	
\begin{figure}[htbp]
			\centering
			\includegraphics[width=0.85\columnwidth]{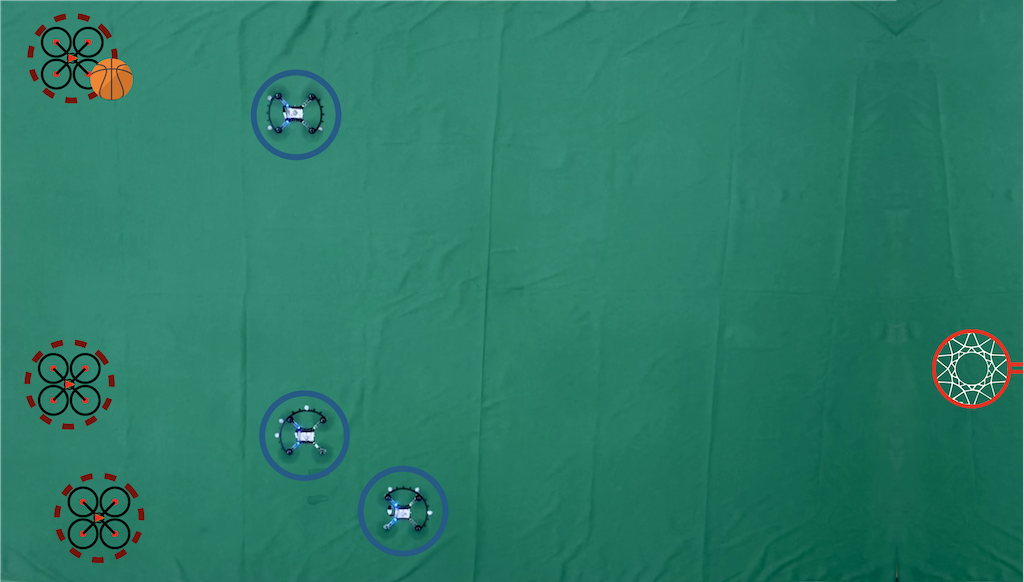}
			
			\vspace{1mm}
			\includegraphics[width=0.85\columnwidth]{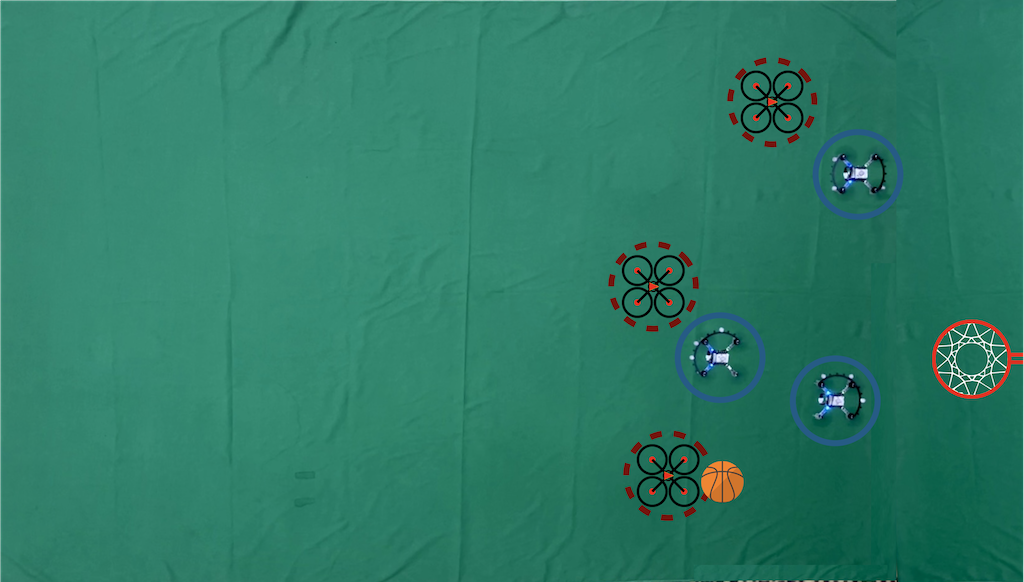}
			
			\caption{Snapshot from an experiment. Real defenders are highlighted by blue circles and
				virtual attackers are depicted with red quadrotors.
			}
			\label{fig:snapshot_exp}
		\end{figure}

	In this subsection, we consider a multi-robot scenario related to a basketball game. The scenario concerns a team of $N = 3$ robots that play the role of the defending team, while the offending team, having also $3$ players, pursues pre-defined trajectories. 
	Each defending player $i$ is associated to an offender with location $p_{i}^t \in \real^3$ at time $t$. Moreover, the defending team needs to take into account also the position of the ball at time $t$ denoted by $b^t \in \real^3$. The strategy of each defending player $i$ consists in placing itself as close as possible to a point lying on the segment linking the location of the basket (denoted as $p_{\text{bsk}} \in \real^3$) with the one of the $i^{\text{th}}$ offending player. Moreover, the defending team also aims to maintain its center of mass as close as possible to a point lying on the segment linking the basket and the ball. Finally, the robots also want to avoid collisions among them. Such a defending strategy is captured by an instance of problem~\eqref{eq:aggregative_problem} with objective functions
	\begin{align*}
		\fag[i][t](\xag[i],\sigma(\xag),\xjneigh)&= \gamma_{i,p}\norm{x_i - \tilde{p}_i^t}^2 + \gamma_{\text{agg}}\norm{\sigma - \tilde{b}^t}^2 
		\notag\\
		&\hspace{.5cm}
		+ \sum_{j\in \innbrs_i^t} -\log(\Vert\xag[i]-\xag[j]\Vert),
	\end{align*}
	where $\lambda_i, \lambda_{\text{agg}} \in [0,1]$. Similarly to Section~\ref{sec:surveillance_strategies}, $\tilde{p}_i^t = \lambda_i p_{\text{bsk}} + (1-\lambda_i)p_i^t$ models a point lying on the segment between the basket and the $i^{\text{th}}$ offensive player, while $\tilde{b}^t = (1-\lambda_{\text{agg}})p_{\text{bsk}} + \lambda_{\text{agg}}b^t$ models a point lying on the segment between the basket and the ball. 
	Similarly to the description in Section~\ref{sec:task3}, the feasible set of each player $i \in \agents$ is
	\begin{align*}
		\begin{cases}
			[p_{i}^t]_c + \epsilon_c \le [\xag[i][t]]_c \le [p_{\text{bsk}}]_c  & \text{if } [p_{i}^t]_c \le [p_{\text{bsk}}]_c 
			\\
			[p_{\text{bsk}}]_c \le [\xag[i][t]]_c \le [p_{i}^t]_c  - \epsilon_c & \text{if } [p_{i}^t]_c > [p_{\text{bsk}}]_c 
		\end{cases} \quad \!\!\!c =1, 2, 3,
	\end{align*}
	$\epsilon_c^t = \max(\epsilon_{c,\text{min}},\kappa_c |[p_{i}^t]_c - [p_{\text{bsk}}]_c  |^2)$ with $\kappa_c, \epsilon_{c,\text{min}} > 0$.
	To simulate a realistic behavior,  the offending team follows pre-defined trajectories, and at a certain point, the ball is passed between two teammates. 
	In the proposed experiments, we choose a robotic platform for the Crazyflie~$2.0$ nanoquadrotor. Intruders are virtual, simulated quadrotors. A snapshot from an experiment is in Figure~\ref{fig:snapshot_exp}, while a video is available as accompanying material to the 
	paper\footnote{The video is also available at \url{https://www.youtube.com/watch?v=5bFFdURhTYs}}.

	\section{Conclusion}
	\label{sec:Conclusions}
	
	This work proposed a distributed algorithm to control a team of cooperating robots protecting a target from a set of intruders. The strategy of the defending team has been modeled according to a distributed online aggregative optimization framework.
	Specifically, the robotic team determines its configuration in the space according to
	the intruder positions, team barycenter position with respect to the target, and inter-robot collisions.
	We corroborated the effectiveness of our method with simulations and experiments on a team of cooperating quadrotors.
	
	\balance
	
	\bibliographystyle{IEEEtran}
	\bibliography{biblio_aggregativematches}
	
\end{document}